\documentclass[a4paper,fleqn]{cas-dc}

\usepackage[numbers]{natbib}
\usepackage{subfig}
\usepackage{siunitx}
\usepackage{tabularx}
\usepackage[inline]{enumitem}

\def\tsc#1{\csdef{#1}{\textsc{\lowercase{#1}}\xspace}}
\tsc{WGM}
\tsc{QE}

\begin{document}
\let\WriteBookmarks\relax
\def\floatpagepagefraction{1}
\def\textpagefraction{.001}

\shorttitle{}    

\shortauthors{}  

\title [mode = title]{Improving dependability in robotized bolting operations}  

\author[1]{Lorenzo Pagliara}[orcid=0000-0001-8283-0320]
\ead{lpagliara@unisa.it}
\credit{Conceptualization, Data curation, Formal analysis, Investigation, Methodology, Software, Validation, Visualization, Writing – original draft, Writing – review \& editing}

\author[1]{Vincenzo Petrone}[orcid=0000-0003-4777-1761]
\ead{vipetrone@unisa.it}
\credit{Writing – review \& editing}

\author[2, 3]{Violeta Redondo}[orcid=0000-0003-0238-7280]
\ead{violeta.redondo.gallego@upm.es}
\credit{Formal analysis, Investigation, Methodology, Writing – review \& editing}

\author[1]{Enrico Ferrentino}[orcid=0000-0003-0768-8541]
\ead{eferrentino@unisa.it}
\cormark[1]
\credit{Conceptualization, Formal analysis, Investigation, Writing – original draft, Writing – review \& editing}

\author[3]{Manuel Ferre}[orcid=0000-0003-0030-1551]
\ead{m.ferre@upm.es}
\credit{Conceptualization, Formal analysis, Investigation, Resources, Supervision, Writing – review \& editing}

\author[1]{Pasquale Chiacchio}[orcid=0000-0003-3385-8866]
\ead{pchiacchio@unisa.it}
\credit{Conceptualization, Resources, Supervision, Writing – review \& editing}

\affiliation[1]{organization={Department of Information Engineering, Electrical Engineering and Applied Mathematics (DIEM), University of Salerno},
            addressline={ Via Giovanni Paolo II, 132}, 
            city={Fisciano},
            postcode={84084}, 
            state={SA},
            country={Italy}}

\affiliation[2]{organization={GTD Science, Infrastructures \& Robotics},
            addressline={Av. Leonardo Da Vinci, 2}, 
            city={Madrid},
            postcode={28906}, 
            country={Spain}}

\affiliation[3]{organization={Universidad Politécnica de Madrid, Centre for Automation and Robotics (CAR) UPM-CSIC},
            addressline={José Gutiérrez Abascal, 2}, 
            city={Madrid},
            postcode={28006}, 
            country={Spain}}

\cortext[1]{Corresponding author}



\begin{abstract}
Bolting operations are critical in industrial assembly and in the maintenance of scientific facilities, requiring high precision and robustness to faults. Although robotic solutions have the potential to improve operational safety and effectiveness, current systems still lack reliable autonomy and fault management capabilities. 
To address this gap, we propose a control framework for dependable robotized bolting tasks and instantiate it on a specific robotic system. 
The system features a control architecture ensuring accurate driving torque control and active compliance throughout the entire operation, enabling safe interaction even under fault conditions.
By designing a multimodal human-robot interface (HRI) providing real-time visualization of relevant system information and supporting seamless transitions between automatic and manual control, we improve operator situation awareness and fault detection capabilities. 
A high-level supervisor (SV) coordinates the execution and manages transitions between control modes, ensuring consistency with the supervisory control (SVC) paradigm, while preserving the human operator's authority.
The system is validated in a representative bolting operation involving pipe flange joining, under several fault conditions.
The results demonstrate improved fault detection capabilities, enhanced operator situational awareness, and accurate and compliant execution of the bolting operation.
\end{abstract}





\maketitle

\section{Introduction}\label{sec:introduction}
Bolting plays a vital role in both industrial assembly processes \cite{jia_survey_2019} and in the maintenance of large scientific facilities \cite{redondo_design_2025}.
In scientific infrastructures, bolts and screws are widely employed across several systems requiring periodic maintenance.
Threaded fasteners are commonly used to mount components to their supports \cite{arranz_remote_2020}, and precise alignment is often achieved by adjusting multiple bolts \cite{arranz_alignment_2024}.
To avoid the need for cutting or welding, pipe flanges are typically joined using chain clamps --- in which the nut is captive \cite{redondo_design_2025} ---, allowing for the replacement of pipe sections without requiring a second support device to hold the fastener \cite{chen_spallation_2001}.
In industrial manufacturing, it is estimated that $37.9\%$ of all mechanical assembly tasks involve bolt insertion and tightening \cite{martin-vega_industrial_1995}.

\begin{figure}
    \centering
    \subfloat[Coupling \label{fig:coupling}]{\includegraphics[width=0.5\columnwidth]{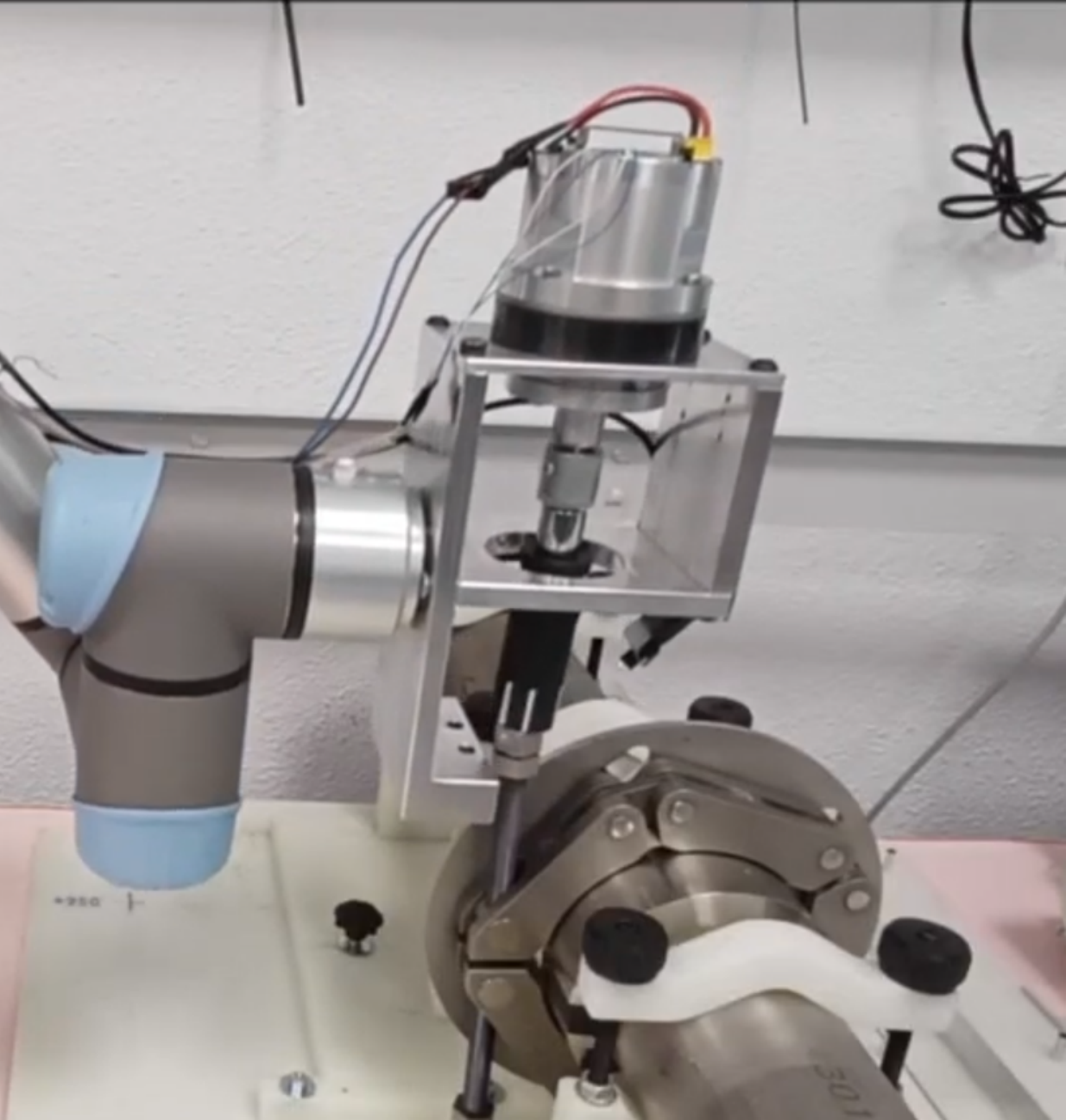}}
    \hfill
     \subfloat[Tightening \label{fig:tightening}]{\includegraphics[width=0.5\columnwidth]{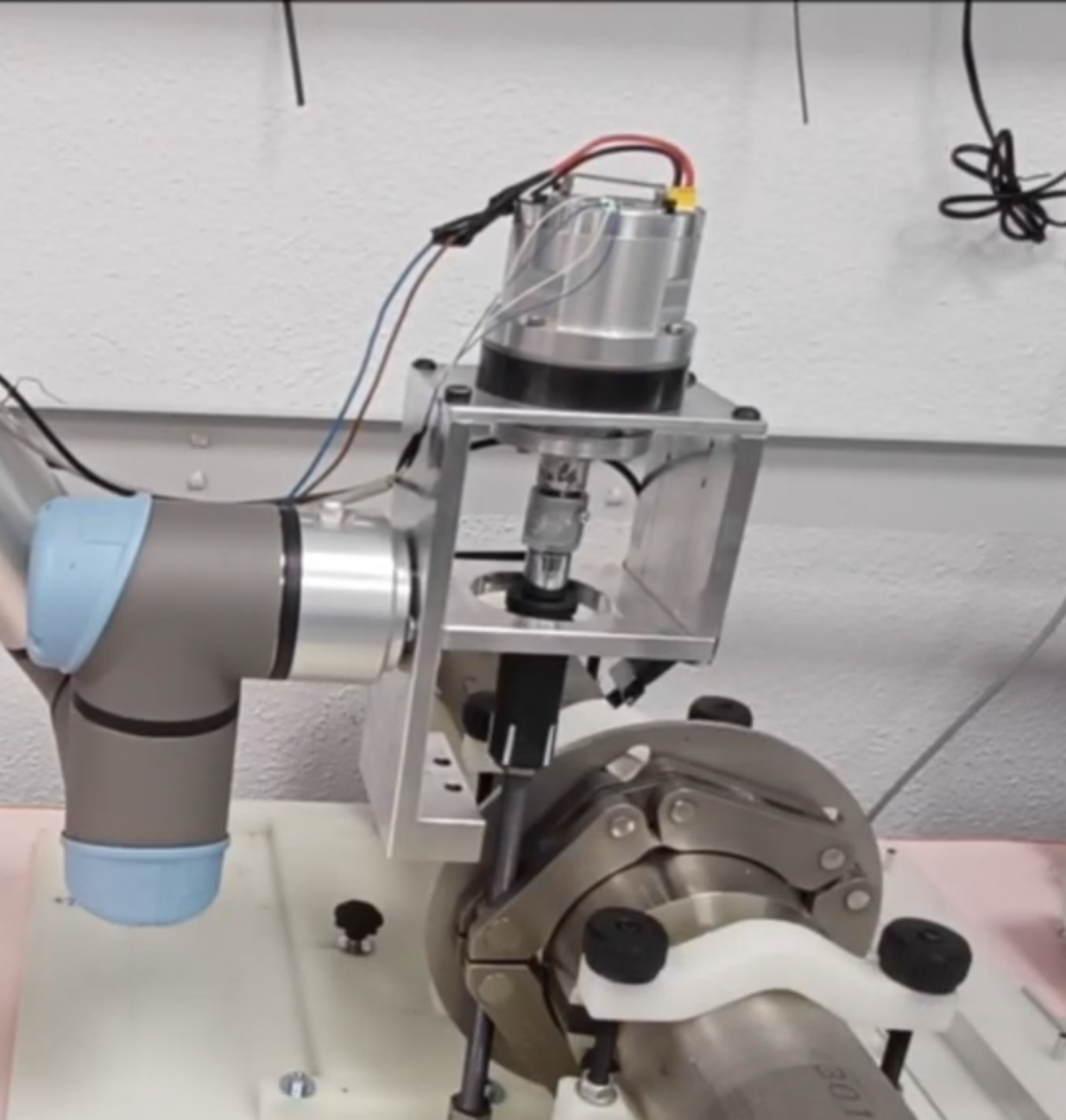}}
    \caption{Example of coupling and tightening steps in a bolting operation.}
    \label{fig:coupling-and-tightening}
\end{figure}

Despite their prevalence, bolting operations remain among the most challenging tasks to fully automate \cite{jia_survey_2019}, particularly in high-mix, low-volume production settings and in scientific maintenance, where tasks vary frequently and dedicated fixtures or precise part positioning are often unavailable.
Indeed, many bolting operations are currently performed either manually by human operators, or through repetitive automated systems that rely on complex fixtures and dedicated part feeders \cite{dharmara_robotic_2018}. 
Manual bolting offers flexibility in handling different tools and adaptability to variations in geometry and positioning.
However, it requires the operator to be physically present at the task site, hence being labor-intensive, and is prone to procedural errors even when a calibrated torque wrench or automatic tool is used.
In a recent study concerning manually tightening a sequence of 5 bolts with a calibrated torque wrench, \SI{67.6}{\percent} of participants made mistakes, e.g., incorrect tightening order, misidentified screw locations, or failure to update torque settings \cite{Fau_2025}.
Conversely, repetitive automation ensures higher efficiency and accuracy, but it offers limited flexibility and poor adaptability to environment variations.
This highlights the demand for an automation approach that combines the flexibility and adaptability of manual execution with the skills of automatic systems.
The literature proposes a variety of automation approaches \cite{dhayagude_fuzzy_1996}, including 2D and 3D vision-guided methods.
Early approaches \cite{pitipong_automated_2010} rely on 2D vision to monitor fastener alignment, but their accuracy is severely affected by environmental conditions and surface reflectivity. 
Hybrid sensing strategies, combining 2D visual feedback with force/torque feedback \cite{chen_integrated_2007, chen_high-precision_2009}, improve insertion accuracy and enable fixtureless assembly, in some cases supported by learning-based correction mechanisms \cite{penacabrera_machine_2005}. 
Despite these advances, 2D vision–based methods remain limited in degrees of freedom and prove effective only in structured environments.
To address these limitations, 3D vision systems are increasingly adopted for fixtureless operations \cite{diaz_robotic_2025}, enabling the direct detection of spatial features from component geometry \cite{bone_vision-guided_2003}.
Solutions based on rapid object registration techniques aim at maximizing robustness and computational efficiency, albeit at the expense of accuracy \cite{bohnke_fast_2010}.
Geometric feature–based recognition methods \cite{lee_3d_2012} ensure high accuracy but require continuous visibility of reference features, while geometric primitive fitting techniques involve longer processing times and significant pose estimation errors \cite{oh_stereo_2012}.
Building on these principles, surface matching techniques are employed for fixtureless bolting operations \cite{ogun_3d_2015}. An ongoing challenge concerns the detection of holes on multiple planes, especially in the presence of internal threads. Existing methods \cite{wang_effective_2012, bendels_detecting_2006} rely respectively on relationships between planes and edges, unsuitable for non-solid holes or on point set analysis, which requires several minutes to process medium-resolution point clouds.
Therefore, although recent research continuously improves the integration of vision, force feedback, and learning-based strategies, current autonomous robotic systems for bolting operation still face limitations in terms of real-time performance, accuracy, reliability and adaptability to unstructured environments \cite{chen_automated_2021}.

Furthermore, sporadic fault conditions may arise during bolting operations, which can lead to critical failures if undetected \cite{myer_ixtoc_1984}.
Typical fault conditions include part misalignment, over-tightening, cross-threading, and jamming \cite{jia_survey_2019}.
Cross-threading occurs when the external thread engages the internal one incorrectly, causing the threads on one side not to follow the same spiral as those on the opposite side, while jamming can occur when the tilt-angle control is too rigid and the initial inclination exceeds the locking angle \cite{nicolson_compliant_1993}.
In some cases, hardware failures can occur in the power tools used to perform bolting operations. For this reason, some studies investigate the use of robotic systems to operate hand tools originally designed for manual use \cite{tang_robotic_2023}.
Several works emphasize the use of wrench signals as the primary sensing modality for fault detection \cite{klingajay_comparison_2003, wilson_fastener_2012, matsuno_fault_2013}.
However, these methods typically cover only a narrow range of failure modes.
Learning-based techniques have shown promise, particularly in structured low-variability environments, but often struggle to generalize under real-world uncertainty \cite{althoefer_automated_2008, moreira_online_2018}.
Similar challenges arise in recent data-driven approaches to fault detection \cite{wende_ml-pipeline_2024, yang_machine-learning-enabled_2024}.

Finally, compliance has long been acknowledged as a critical factor in autonomous bolting, to the extent that nearly all commercially available tools integrate compliant mechanisms in their design \cite{jia_survey_2019}. 
In addition, active compliance techniques, such as impedance and force control, are critical to improve safety and dependability during operations \cite{manyar_physics-informed_2023}, especially when compared with rigid motion controllers
 \cite{claudia_carvalho_robot_2024}.

In large scientific facilities, such as nuclear plants, unreliable autonomy or missed fault detection can have severe consequences.
In such high-stakes contexts, telerobotics offers a powerful alternative. By extending human sensing and manipulation capabilities into remote, hazardous, and unstructured environments, it enables the safe execution of complex tasks and ensures immediate and effective detection and response to faults \cite{bicker_early_2004}.

A pioneering example is the Joint European Torus (JET), where direct teleoperation was successfully employed in 1998 to replace the facility's divertor \cite{rolfe_perspective_2007}. Since then, telerobotics has become an essential component of maintenance strategies for major scientific infrastructures \cite{buckingham_remote-handling_2016}. In some instances, it has been integrated only after the plant's construction, with maintenance procedures progressively adapted for robotic execution. Several examples of this approach can be found at CERN, where the MIRA robot is currently used to perform radiation surveys in the tunnel \cite{forkel_towards_2024}, alongside many other teleoperated interventions. 

Nowadays, increasing efforts are being made toward achieving higher levels of automation in selected tasks, enabling faster and safer maintenance operations \cite{buckingham_remote-handling_2016}. In such scenarios, the most commonly adopted paradigm is supervisory control (SVC) \cite{sheridan_telerobotics_1992}, where specific phases of the procedure are automated, while the human operator retains control over transitions and can take control of the system at any time.
Furthermore, new facilities under development are being designed with telerobotic compatibility in mind from the outset. In these cases, maintenance requirements are embedded directly into component design, and dedicated robotic interfaces are implemented to ensure seamless integration with telerobotic systems. Notable examples include IFMIF-DONES \cite{micciche_remote_2019}, ITER \cite{lyytinen_development_2021}, and CFETR \cite{zhao_engineering_2020}.

Recent work on human‑in‑the‑loop robotics has explored continuous shared and corrective control, where human and autonomous inputs are blended within a single control mode for tasks such as polishing \cite{Zhou_2025}, aerial inspection \cite{D_Angelo_2025}, and corrective shared autonomy \cite{Hagenow_2021}; other approaches address decision‑level human‑robot hybrid decision‑making \cite{Liu_2026}.
These frameworks are effective for their respective tasks, but they do not address the discrete, supervised transition between autonomous execution and teleoperated recovery for contact‑rich fastening operations --- with explicit switching conditions and an operator validation gate ---, which is the focus of a supervisory control framework \cite{Parasuraman_2000}.

Motivated by the importance of bolting operations in both industrial assembly processes and the maintenance of scientific facilities, and considering the current technological limitations in terms of autonomy and fault detection reliability, in this work we propose a control framework enabling dependable robotized bolting operations, and a robotic system implementing it. 
The framework integrates state‑of‑the‑art components --- active compliance, torque control, multimodal human-robot interface (HRI), and haptic teleoperation --- each of which is individually established in the literature.
Our distinctive contribution lies in the structured organization of the supervised alternation between autonomous and teleoperated execution: the framework formalizes the process as a step‑by‑step pipeline with operator validation gate after each step, ensuring consistency with the supervisory control paradigm while preserving the human operator's ultimate authority.
A high‑level supervisor (SV) orchestrates the components and manages the transitions between control modes, allowing for a seamless transition from autonomous to manual operations, and vice versa.

Given the practical impact of the proposed approach, a key contribution is a TRL-5 experimental validation of the system on a representative bolting operation, consisting of bolt tightening to join pipe flanges, under several fault conditions:
\begin{enumerate}
    \item control system failure in the absence of active compliance under part misalignment;
    \item bolt pose estimation failure;\label{itm:vision-sytem-failure}
    \item hardware failure of the bolt driver used to complete the tightening process.
\end{enumerate}
Notably, we validate the framework in condition \ref{itm:vision-sytem-failure} \textit{assuming} the availability of a vision system for nominal task execution and fastener-pose estimation; the design, implementation and anomaly-detection performance of that system are outside the scope of this work and are not validated here.
In the experimental validation, bolt-pose estimation errors are therefore simulated to exercise the contingency-handling mechanism.

\section{Methodology}\label{sec:methodology}
Robotized bolting operations involve multiple steps, often including feeding, alignment, tightening, and postfastening, each relying on specific components and control strategies.
The success of each step is essential to enable the correct execution of the following ones.
In the most general case of an unstructured environment, the bolting task involves six degrees of freedom (DoF): five define the bolt alignment with the parts to be joined, or equivalently, the driver unit alignment with a fixed bolt, in terms of position in the three-dimensional space and axis orientation; the other one corresponds to the bolt's axial rotation and tightening.
This latter DoF is typically provided by an independent actuator integrated into the bolt driver unit, which is designed for high-speed rotation and high-stiffness tightening.
\begin{figure*}
    \centering
    \includegraphics[width=.9\textwidth]{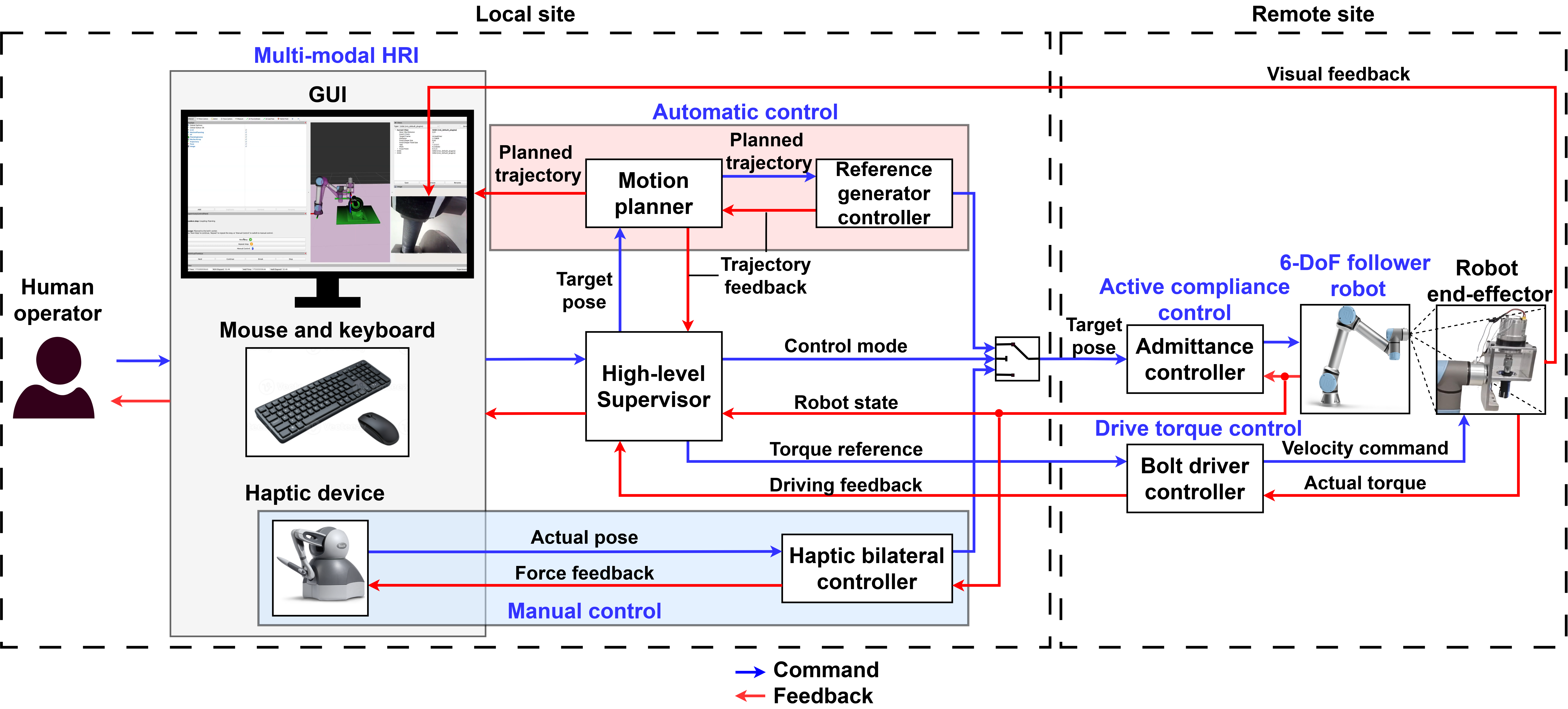}
    \caption{Control system for dependable bolting operations.}
    \label{fig:SVCS}
\end{figure*}

The dependability of robotized bolting operations can be compromised by several potential failures in sensors, control and actuation systems, which may lead to undesired behaviors.
Given the critical nature of bolting tasks, dependability must be ensured not only at the component level but also across the entire robotic system.
Dependability is an integrated concept that encompasses a set of essential system attributes, including
robustness and safety.
The former includes availability, reliability, and maintainability \cite{de_santis_atlas_2008}, and refers to the system's capability to fulfill its intended functions in normal operation, but also to maintain acceptable performance under degraded or fault conditions, guaranteeing continuity and predictability of its behavior.
The latter refers to the prevention of catastrophic consequences for the environment or the robot itself, even in the presence of failures, malfunctions or unexpected collisions.
We adopt the dependability taxonomy of \cite{Avizienis_2004}, distinguishing faults, errors, and failures.

In the context of the proposed framework, dependability is pursued through the four means of \cite{Avizienis_2004}: prevention (active compliance avoids excessive forces; torque control prevents over-tightening), tolerance (the supervisor detects faults and transitions to a safe state), removal (manual teleoperation completes the task after a fault), and forecasting (the HRI supports operator prediction of evolving conditions).
The framework improves the dependability attributes of reliability (probability of successful task completion) and safety (avoidance of damage to equipment or environment) for the fault classes considered.

One major challenge of robotized bolting operations in unstructured environments lies in relying on reliable sensing and actuation systems.
As discussed in Section~\ref{sec:introduction}, unreliable sensing, perception, and actuation can significantly affect the robot's ability to complete the operation effectively, as inaccuracies in environment perception or torque application may compromise both the proper execution of the task and the overall system dependability.
This becomes particularly critical considering that such unreliability constitutes the primary source of faults, but also leads to failures in detecting them.
It is therefore useful to explicitly define the types of faults that can compromise the operation: internal faults and interaction faults.
The former include damage to mechanical components and actuators and/or sensor faults, such as driver unit failures.
The latter include issues related to interaction with the environment, such as parts misalignment that cause the system to exceed its nominal limits.

Achieving dependability requires a set of strategies aimed at preventing, detecting, isolating, and actively handling such faults.
This calls for modular human-in-the-loop control architecture and multimodal HRI, which facilitate the monitoring, detection and handling of potential faults, and ensure the correct execution of each step of the operation.
In addition, dependability can be achieved by exploiting the intrinsic redundancy of the robotic system with respect to the task, as most robots used in industrial and research environments are 6-DoF and therefore inherently redundant for bolting operations.
Indeed, this extra DoF can be operatively employed to achieve bolting even in the case of a failure of the driver unit.

Regarding safety, Section~\ref{sec:introduction} discussed how it is closely related to compliance and to the achievement of the desired driving torque, in order to prevent joint component failures.
This requires active compliance techniques, as well as accurate control and monitoring of the driving torque and alignment angle, to also avoid common faults such as cross-threading and jamming.

Building upon this analysis, we propose a control framework for dependable bolting operation integrating monitoring and control components, each designed to fulfill specific requirements, and supporting autonomous and manual operations.
Such framework integrates:
\begin{itemize}
    \item \textit{6-DoF follower robot (FR)}: intrinsically redundant with respect to the bolting task and equipped with all the tools, i.e., driver unit, force/torque sensor and cameras, required to perform the operation;
    \item \textit{active compliance control}: ensuring precise alignment angle control and guaranteeing operational safety;
    \item \textit{driving torque control}: ensuring that the desired clamping force is achieved while providing operational safety and effectiveness 
    \item \textit{automatic control:} translating a target pose, nominally supplied by a vision system, into a collision-aware trajectory for autonomous execution of the task;
    \item \textit{manual control}: including hardware interfaces and controllers to enable manual intervention and fault handling;
    \item \textit{multimodal HRI}: providing a unified environment for the supervision of operation execution, facilitating fault detection and handling on the part of a human operator;
    \item \textit{high-level supervision}: orchestrating different components and allowing for a seamless transition between autonomous and manual control.
\end{itemize}
With reference to Figure~\ref{fig:SVCS}, which illustrates the system implementing the proposed control framework, the following sections describe its individual sub-components.
In Section~\ref{sec:multimodal-hri} a detailed description of the multimodal HRI is provided. 
Section~\ref{sec:active-compliance-control} and Section~\ref{sec:driving-torque-control} present, respectively, the admittance controller used to achieve active compliance and the Bolt Driver controller (BDC) used to reach the desired driving torque.
Finally, in Section~\ref{sec:manual-control} the description of the direct teleoperation system used for manual fault handling, is provided, while in Section~\ref{sec:high-level-supervision} the objectives and the operation of the SV are presented.
The vision system that supplies the automatic control component with the target pose is not designed or implemented as part of this work: it is emulated in the system instantiating the proposed framework, and the bolt pose is therefore assumed to be known throughout the experimental validation.
Nevertheless, the proposed framework is compatible with existing robust detection-based algorithms, such as the one described in \cite{espinosa_peralta_robust_2023}, which could be integrated to achieve a higher level of end-to-end fidelity in future implementations.

The framework's applicability envelope and its formalization-level scope are discussed in Section~\ref{sec:scope-and-applicability}, after the system's components are presented in Sections~\ref{sec:multimodal-hri}--\ref{sec:high-level-supervision}; the implementation- and hardware-specific limitations of the current demonstrator are discussed in Section~\ref{sec:limitations}, and the way the framework improves dependability is discussed throughout Section~\ref{sec:discussion}, supported by an overview of the specific fault classes we validate and the results collected in our experimental campaign, both reported in Section~\ref{sec:experimental-validation}.

\subsection{Multimodal human-robot interface}\label{sec:multimodal-hri}
In robotized bolting operations, the operator's ability to maintain a high level of situational awareness is essential for a safe and effective operation. 
Situational awareness encompasses the perception of relevant elements in the environment, the understanding of their significance, and the prediction of how they may evolve over time. 
Therefore, a suitable design of the HRI plays a fundamental role in supporting situational awareness, as it serves as the primary communication channel between the human and the robotic system.

In this work, we propose a multimodal HRI providing a comfortable and uniform environment for the control of the FR.
This includes standard input devices, such as a keyboard and mouse, along with a haptic device (HD) for direct teleoperation, and a Graphical User Interface (GUI) that communicates bidirectionally with the SV.

\begin{figure*}
    \centering
    \includegraphics[width=\textwidth]{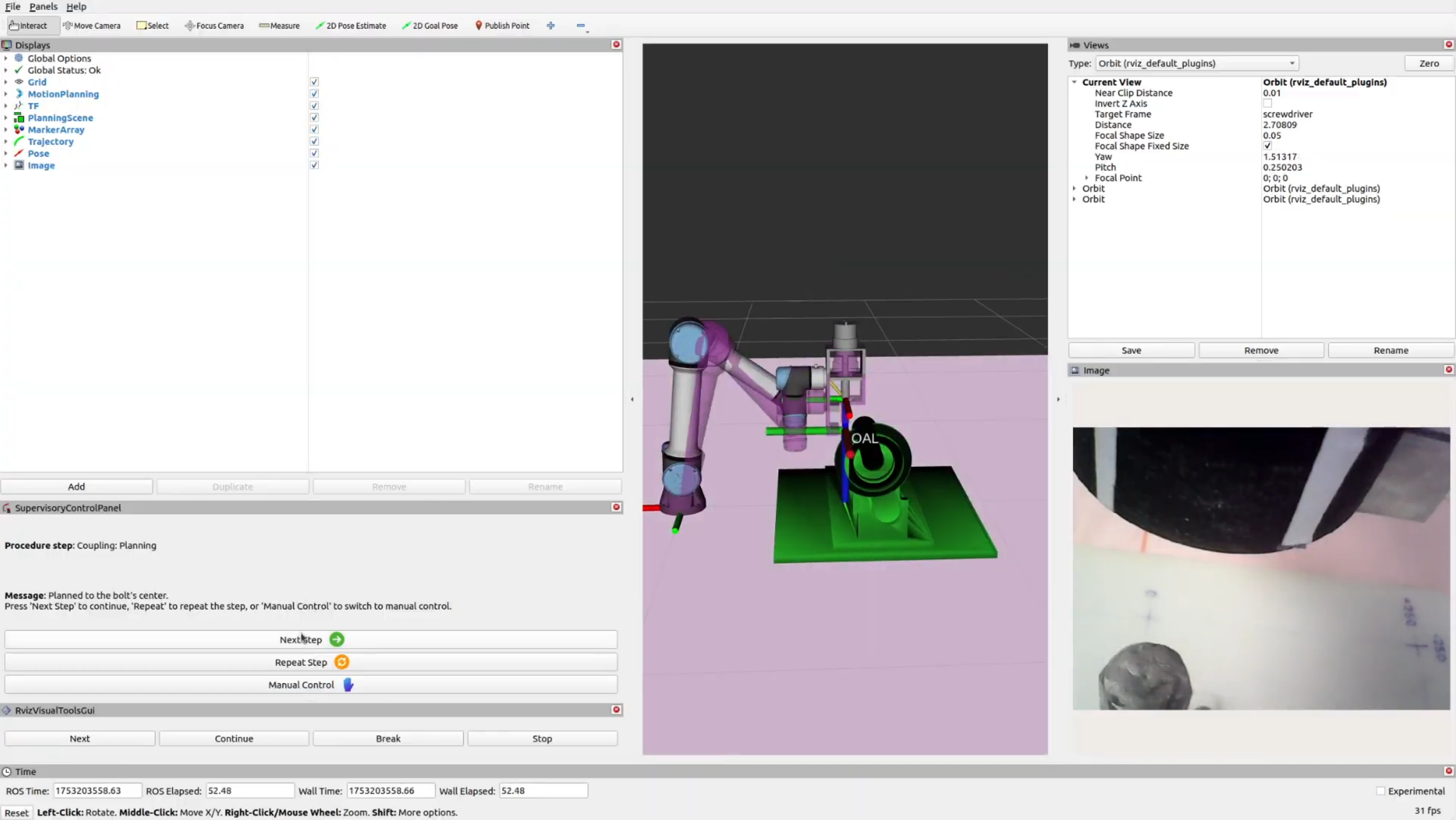}
    \caption{Screenshot of the system's GUI during operation.}
    \label{fig:svcs_gui}
\end{figure*}

The GUI, shown in Figure~\ref{fig:svcs_gui}, is designed to support the human operator in detecting fault conditions. 
To this end, it is designed around six main objectives:
\begin{enumerate}
    \item Monitor the FR's state in real time, including its joint configuration, through continuous updates visualized in a 3D environment. This allows the operator to evaluate the results of the sent commands.
    \item Visualize a 3D model of the interaction environment to support validation by the human operator.
    \item Plan, rehearse, and validate collision-aware trajectories, as done in \cite{pagliara_human-robot_2023}, during approach, coupling, and distancing steps. Eventually, the planned trajectories are transmitted to the robot. 
    \item Manage the transitions between automatic control and manual control, i.e., support the operator in engaging the FR with the HD when the orientation of the stylus and that of the bolt driver tip match.
    \item Monitor the progress of the procedure in real time through the video stream from the camera mounted on the end-effector of the FR.
    \item Enable the human operator to monitor the operation status through dedicated panels displaying state information and buttons for sending commands to the FR.
\end{enumerate}
Alongside the GUI, real-time visualization of interaction forces and tightening torques is provided through the PlotJuggler tool.
Such visualization enhances the operator's situational awareness and facilitates verification of the operation completion.

\subsection{Active compliance control}\label{sec:active-compliance-control}
In bolting operations ensuring active compliance during interaction is recognized as a fundamental requirement, especially to prevent failure conditions such as jamming.

Therefore, in our work, we adopt an admittance controller as the low-level controller for the FR.
This controller has three key features. 
First, it guarantees the stability of the FR even in the event of communication loss, as it continues to follow the last received reference in the absence of new commands.
Second, it prevents damage to both the FR and the manipulated environment by adapting the received references to enable compliant interaction along both translational and rotational axes. 
This also facilitates the bolt alignment.
Finally, it ensures command continuity by discarding reference poses that are too distant from the robot's current state, based on a configurable threshold.
These features make the admittance controller particularly suitable for contact-rich tasks like bolting, where dependability and responsiveness to dynamic interactions are essential.

When the system operates under automatic control, the procedure is executed by following collision-aware trajectories. These are translated into references for the admittance controller by a Reference Generator Controller (RGC). At each control cycle, it provides a pose reference to the admittance controller according to the timing law of the input trajectory.  

This modular structure ensures compatibility between trajectory execution and low-level interaction control.  
Moreover, it allows for flexible switching between control modes without modifying the underlying control architecture.

\subsection{Driving torque control}\label{sec:driving-torque-control}
In both industrial assembly processes and the maintenance of large scientific facilities, ensuring secure tightening of bolts used to assemble different systems is essential to prevent catastrophic failures.

For this reason, in this work we integrate a bolt driver in the FR's end-effector and design a BDC that secures the bolt with the appropriate driving torque.  
This controller receives the target torque and commands a constant velocity to the bolt driver until the measured torque reaches the desired value.  
It also includes a safety mechanism that allows the tightening process to be interrupted upon reception of a stop signal. 

The BDC can also receive position references, enabling it to rotate the bolt driver by a specified amount to safely complete the bolt releasing operation.

These functionalities make the BDC a reliable and flexible component for managing torque-controlled bolting operations, ensuring both operational safety and adaptability to task requirements.

\subsection{Manual control}\label{sec:manual-control}
In bolting operations, effective fault detection and management are crucial to prevent catastrophic events. 
To enable manual handling without exposing the human operator to potentially hazardous environments, we enable direct teleoperation of the FR by integrating a Haptic Bilateral Controller (HBC). 
Since the contact-rich nature of such operations requires accurate control of the FR and natural haptic rendering, the HBC adopted in this work is based on the architecture proposed in \cite{pagliara_haptic_2025}, which enables natural teleoperation and precise control of the FR.

Although in this work we refer to the HBC as a single entity, it actually consists of two controllers: a feedforward controller and a feedback controller. 

The former is responsible for coupling the HD's stylus with the octagonal socket of the bolt driver.  
At each control cycle, it retrieves the current pose of the HD, filters it online using a moving average filter to remove human hand tremors, and maps it to a target pose for the FR.  
This mapping ensures motion correspondence between the movements performed by the human operator with the HD and those executed by the FR in the camera view.

The feedback controller is responsible for rendering force feedback on the HD, enabling the human operator to perceive the remote interaction.  
The force feedback is obtained by a second impedance model, which leverages the motion error generated by admittance control to provide a more comfortable and natural experience for the operator \cite{pagliara_safe_2024}.  
To this end, the feedback is also coordinated with the interaction observed through the camera.

The teleoperation layer's behavior under the network conditions used in the experiments is established in Section~\ref{sec:experimental-setup}; its applicability beyond those conditions, and the delay-compensation techniques compatible with it, are discussed in Section~\ref{sec:limitations}.

\subsection{High-level supervision}\label{sec:high-level-supervision}
The presence of multiple heterogeneous sub-components, each devoted to a specific function, demands a higher-level entity capable of overseeing their operation and ensuring coherent task execution. 
To address this need, we design a SV to monitor and orchestrate the operation and the interaction among the different sub-components. 
By exploiting both the robot's autonomous capabilities and the human operator's supervisory ones, the SV embodies the SVC paradigm, enabling robust, adaptive, and safe operation execution.

To implement this concept, we introduce, in Section~\ref{sec:modeling-svc-operations}, a formal modeling of the SVC operations and define, in Section~\ref{sec:bolt-tightening-pipeline}, a task-specific pipeline that the SV follows during execution, ensuring consistency and dependability.

\subsubsection{Modeling supervisory control operations}\label{sec:modeling-svc-operations}
\begin{figure}
    \centering
    \includegraphics[width=0.8\columnwidth]{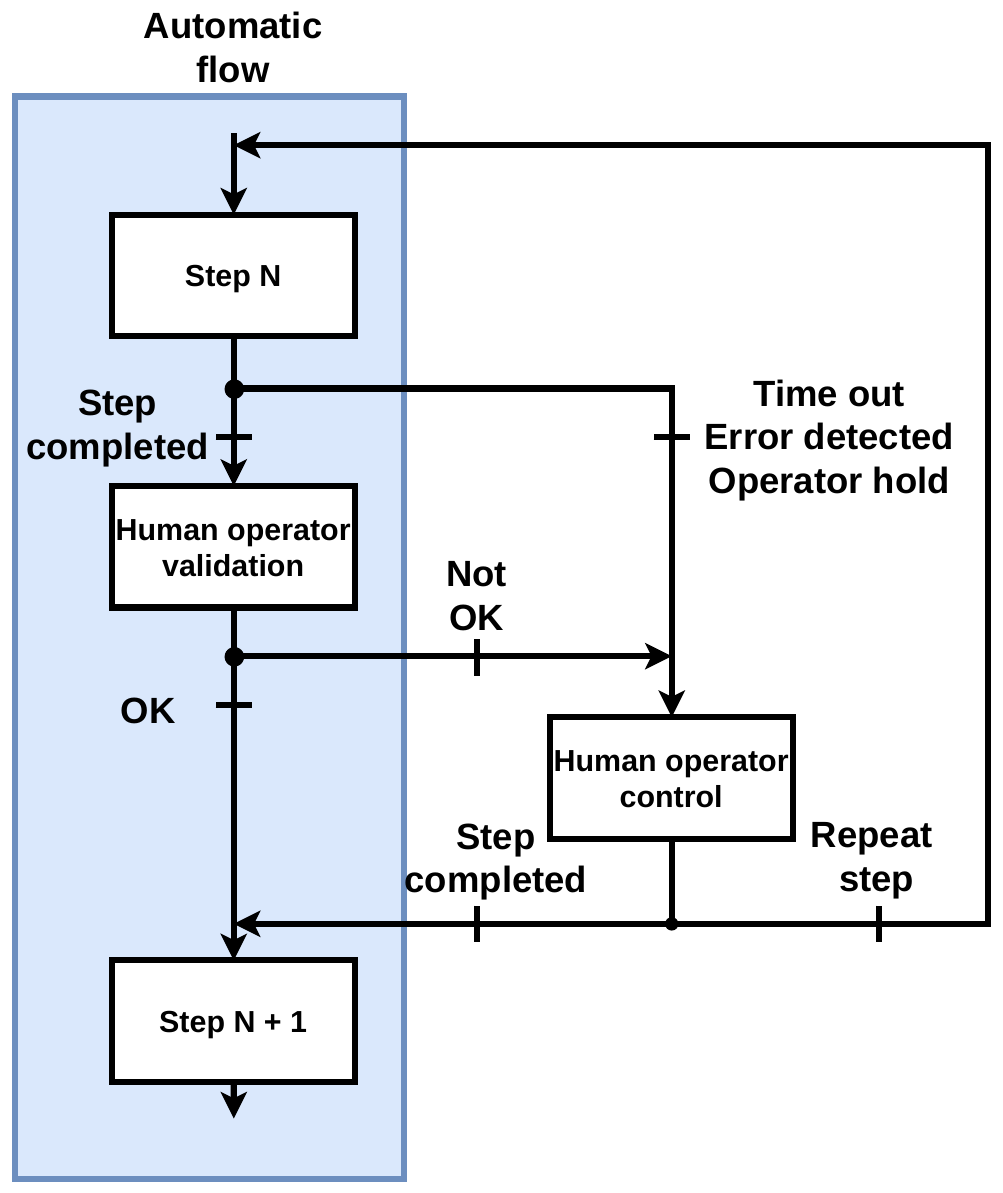}
    \caption{Diagram of a generic telerobotic supervisory control operation.}
    \label{fig:svc_operation_model}
\end{figure}

SVC operations can be effectively modeled as a sequence of discrete steps, each representing a low-level action required to accomplish the overall task.
As shown in Figure~\ref{fig:svc_operation_model}, the system transitions from one step to the next only after receiving validation from the human operator, who retains full authority to intervene at any point, either to repeat a step or to take manual control in the event of a fault. 
Each step concludes with a validation phase, during which feedback is provided to support the operator's decision-making. 

While the system includes basic fault detection capabilities that enable it to recognize potential execution errors and return to the last known safe state when re-execution is required, the primary responsibility for monitoring and handling faults lies with the human operator, who supervises the operation and manages unforeseen situations.

\subsubsection{Bolt tightening pipeline}\label{sec:bolt-tightening-pipeline}
Building upon the general framework presented in Section~\ref{sec:modeling-svc-operations}, in this section we present a task-specific modeling of a tightening operation. The procedure is broken down into the following steps:
\begin{enumerate}
    \item \emph{Approach}: the robot moves in the proximity of the bolt.
    The bolt only needs to be visible in the camera frame, presumably mounted on the end-effector, which is responsible for identifying the bolt's center.
    \item \emph{Bolt identification}: an autonomous vision system identifies the bolt center and provides the estimated pose as a target for the control system.
    In this work, the vision system is emulated (Section~\ref{sec:introduction}).
    \item \emph{Coupling}: the robot precisely moves to the actual bolt location, coupling it with the bolt driver (Figure~\ref{fig:coupling}).
    \item \emph{Tightening}: the bolt driver applies a controlled torque to tighten the bolt according to the specified parameters (Figure~\ref{fig:tightening}). 
    \item \emph{Releasing}: the bolt driver applies a small opposite torque to facilitate the decoupling with the bolt.
    \item \emph{Distancing}: the robot retracts from the bolt location to a safe position, thus completing the operation.
\end{enumerate}
After each step, the operator validates the result based on feedback from the remote site and decides whether to proceed or repeat the step. 
This structured approach allows for safe, flexible, and operator-aware execution of bolting operations in remote and unstructured environments.

\subsubsection{Supervisor}\label{sec:supervisor}
The proposed SV manages the execution of the procedure according to the step-based pipeline defined in Section~\ref{sec:bolt-tightening-pipeline}. 
It supports two control modes, \emph{automatic control} and \emph{manual control}, and is designed to accomplish the following six goals:
\begin{enumerate}
    \item Ensure that transitions between operation steps occur only after execution has been validated by the human operator. 
    This guarantees that the execution flow remains consistent with the SVC paradigm, in which the human operator's authority is always preserved.  
    \item Reset the system to the last known safe state in the event of a fault, whether detected by the system through its basic fault detection capabilities or by the human operator, and allow either the re-execution of the step or the switch to manual control. 
    This feature is critical to ensure the dependability of the execution, as it enables recovery from anomalous conditions without compromising the structural integrity of the system or the surrounding environment.
    \item Interact with the HRI to receive commands and/or validations from the human operator and provide state information of the operation to be displayed on the GUI, to support human decision-making and uphold situational awareness.  
    \item Interface with the planning system to provide reference trajectories to the RGC.
    \item Manage the transition between automatic and manual control, i.e., enable relevant controllers and bring the system to a safe state before switching control.
    \item Provide commands to the BDC to activate the tightening and releasing operations.
\end{enumerate}

\subsection{Scope and applicability}\label{sec:scope-and-applicability}

The proposed framework is a contingency-management and supervisory-control layer --- active compliance, torque control, multimodal HRI, haptic teleoperation, and the logic governing the switch between autonomous and teleoperated execution --- intended to be coupled to an external autonomous bolting executor.

Fault detection is performed by the human operator through the multimodal HRI, consistently with the supervisory control paradigm \cite{sheridan_telerobotics_1992}, in which the operator retains ultimate responsibility for monitoring and intervention.
As already noted above, the perception system is emulated rather than integrated, so we do not claim validated autonomous detection accuracy or latency; the enhanced situational awareness the HRI provides refers to its ability to make relevant state information available to the operator, not to autonomous anomaly detection.

Three specific limitations belong to the formalization itself, independently of the specific hardware instantiated.
The first is that authority is binary and exclusive within a step: the operator either supervises autonomous execution or holds full teleoperated control, and the model does not support continuously blending autonomous and human inputs within a step, e.g., progressively correcting the driving-torque profile or the driver's alignment while a step is executing, rather than stopping and taking full manual control as in corrective or shared-control regimes: this behavior is a requirement of, e.g., collaborative polishing \cite{Zhou_2025} or push-and-slide inspection tasks \cite{D_Angelo_2025}.
The second is that a step's post-condition is bounded by the available instrumentation: validation currently rests on driving torque and operator visual inspection, so a step can be marked complete while the resulting joint remains non-conforming by a criterion not captured by this instrumentation, such as the tests required in practical specifications by facility-level post-maintenance acceptance procedures \cite{Miccich__2025}, which typically depend on factors beyond driving torque alone.
The third concerns manipulator count: the supervisory and HRI layers are formalized around a single FR and a single HD.
Where the fastener's counterpart is not mechanically constrained and must itself be actively held, a bimanual configuration \cite{Choi_2014} would be required; this scenario cannot be accommodated as a drop-in replacement of a single component of our framework, since it requires coordinating two follower arms, correspondingly duplicated haptic interfaces, and expanded camera coverage to frame both interaction points, constituting an extension of the present formalization rather than an upgrade to the instantiated system.

Limitations arising from the specific hardware chosen for the present demonstrator (i.e., a specific instance of our framework) are discussed separately in Section~\ref{sec:limitations}, once the corresponding experimental evidence has been presented in Section~\ref{sec:experimental-validation}.

\section{Experimental validation}\label{sec:experimental-validation}

This section presents an experimental validation of the proposed control framework performed by rehearsing the operation described in Section~\ref{sec:bolt-tightening-pipeline} under different fault conditions of the system implementing it.
The aim of the experiments is to validate the dependability of the system in preventing and handling the most common failure causes.
Three fault classes are exercised, corresponding respectively to Sections~\ref{sec:prevention-of-interaction-faults}, \ref{sec:handling-of-visual-system-failures}, and \ref{sec:handling-of-bolt-driver-failure}: interaction faults (misalignment/jamming), perception faults (incorrect bolt-pose estimates), and hardware faults (driver failure).
An end-to-end demonstration of the system, under nominal conditions, is provided in the accompanying video\footnote{rcim-2025-bolting-operations.mp4}, where the full execution of the operation under automatic control is showcased.

With the aim of confirming the impact of active compliance on preserving the dependability and safety of the system under interaction faults, in Section~\ref{sec:prevention-of-interaction-faults} we compare the execution of the coupling operation under conditions of misalignment between the driver unit and the bolt, both with and without active admittance control.

In order to quantitatively assess the system's effectiveness in assisting the detection and management of faults in the external autonomous system, we simulate a scenario in which the bolt is incorrectly localized, e.g., by a vision system. 
The fault is detected by the human operator only during the execution of the coupling trajectory. 
In response, control is transferred to the operator, who completes the task via direct teleoperation.
The results of this experiment are presented in Section~\ref{sec:handling-of-visual-system-failures}.

Finally, in Section~\ref{sec:handling-of-bolt-driver-failure} we provide a formal assessment of the system's ability to handle hardware failures through the direct teleoperation system, by replicating a bolt unit malfunction scenario, with the human operator performing the tightening step manually.

The experimental campaign reported below is deliberately scoped to isolate the contingency-handling mechanism rather than to explore hardware variants.
It addresses a single fastener size, held in a fixed-socket end-effector; an automatic, vision-based tool-change capability \cite{Blake_2022}, which would let the autonomous executor select among multiple tools by extending the automatic-control component (Figure~\ref{fig:SVCS}) without modifying the supervisory logic, has not been exercised here.
Likewise, the campaign addresses a single-bolt operation: multi-bolt sequences, and the associated preload distribution and tightening order across fasteners \cite{Kumakura_2003}, are compatible with the existing step-based pipeline but have not been exercised in this work.
Finally, we did not vary the assembled parts themselves (e.g., different pipe diameters or joint types): the framework addresses contingency handling during bolting --- the handover between autonomous and teleoperated execution --- which depends on the fastener interaction rather than on the surrounding parts being joined, so this variation would not be expected to alter the reported results.

\subsection{Experimental setup}\label{sec:experimental-setup}

\begin{figure*}
    \centering
    \subfloat[Remote site of the experimental setup, showing the UR5e robotic arm equipped with the mechanical end-effector and the pipes used for experimental validation.
    \label{fig:experimental-setup}]%
    {\includegraphics[width=0.48\textwidth]{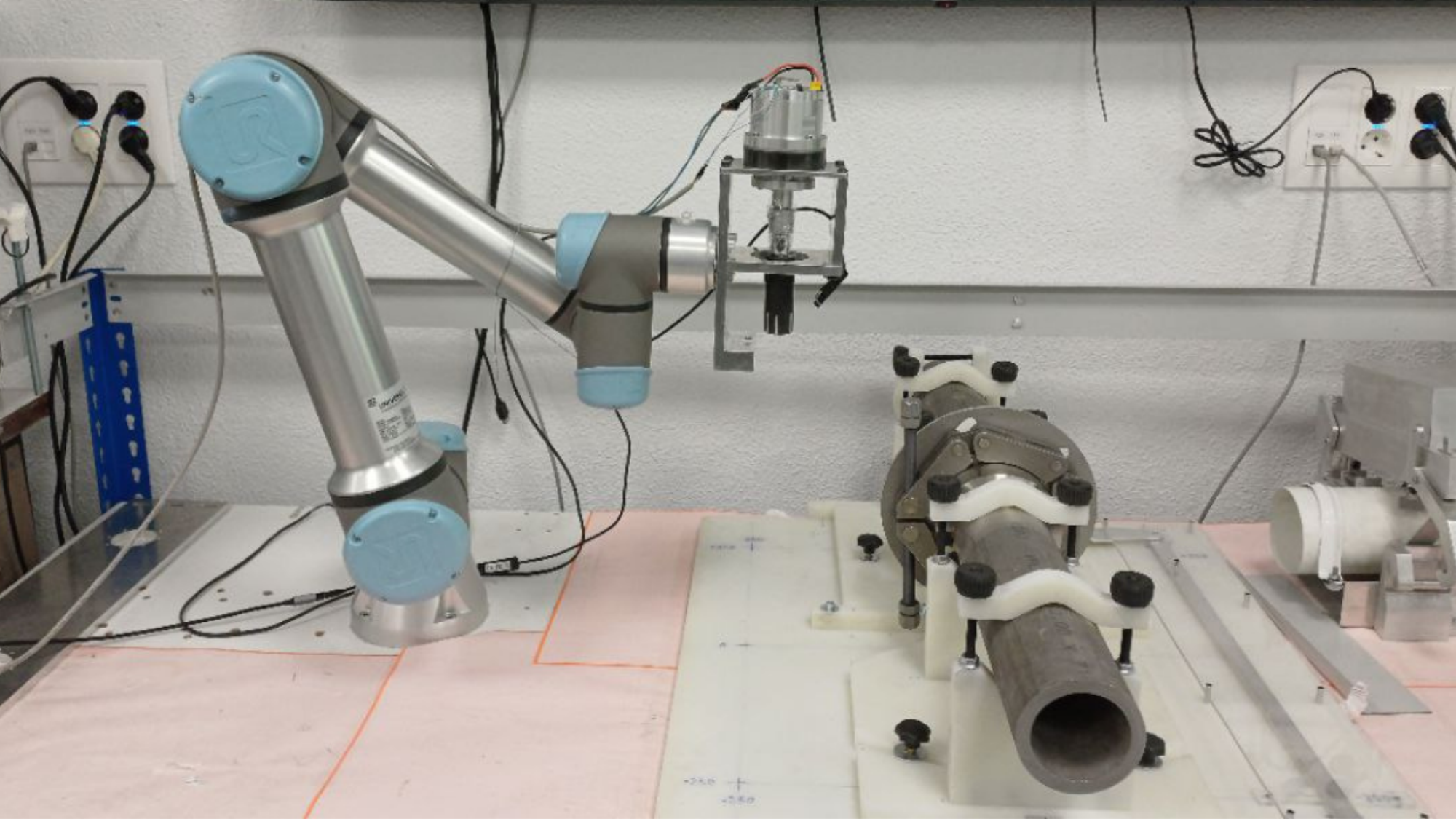}}
    \hfill
    \subfloat[Close-up of the bolted chain clamp used in the experiments. The nut is captive in the clamp link (highlighted), and the robot engages the bolt head; the clamp itself closes the reaction path.
    \label{fig:nut-bolt-closeup}]%
    {\includegraphics[width=0.48\textwidth]{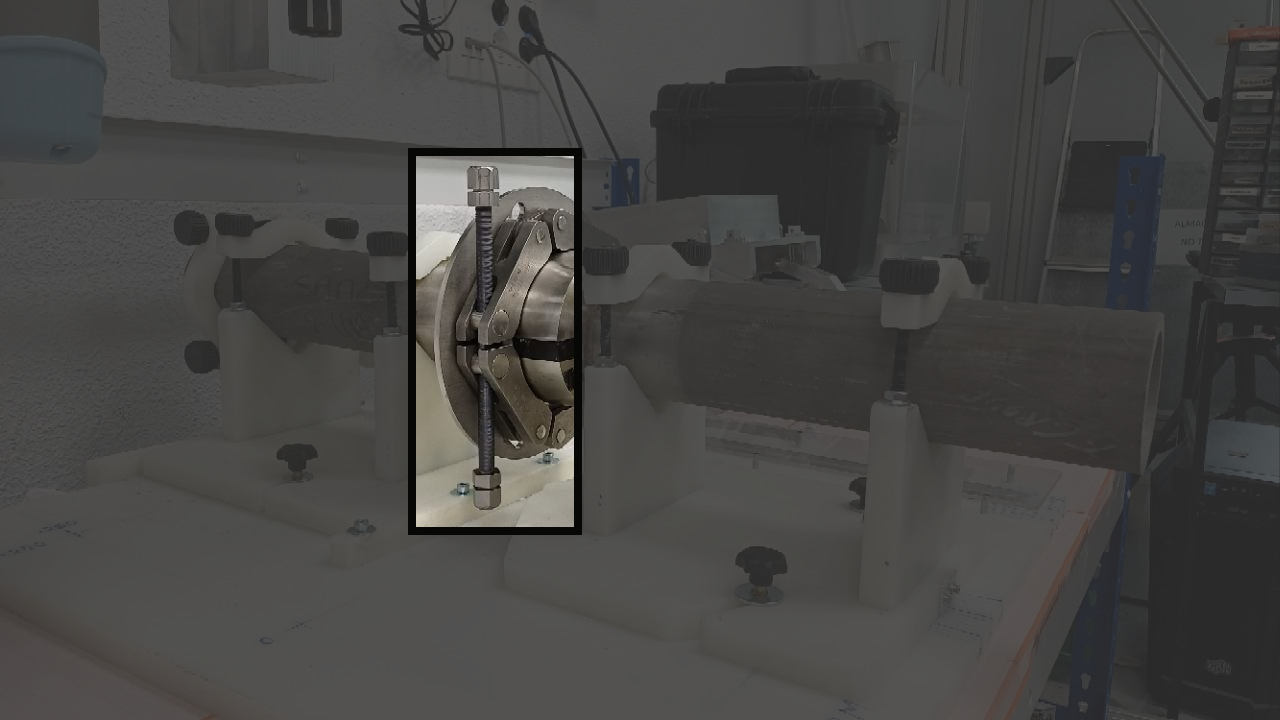}}
    \caption{Experimental setup for the bolting operation: (a) overall remote site with the UR5e robot and the pipe assembly; (b) detailed view of the chain clamp and the captive nut arrangement.}
    \label{fig:experimental-setup-combined}
\end{figure*}

The experimental setup involves the following equipment and materials:
\begin{itemize}
    \item a workstation running Ubuntu 22.04 with ROS2 Humble. The latter is a general purpose computer communicating with the FR over an Ethernet network. To guarantee deterministic and reliable control execution, the operating system was enhanced with the PREEMPT\_RT real-time patch \cite{Reghenzani_2019};
    \item a 6-DoF force-feedback HD \cite{3d_systems_touch_2016};
    \item a 6-DoF UR5e robotics arm by Universal Robots \cite{universal_robots_2025}. It is equipped with a mechanical end-effector for bolting operations, featuring a camera for the visual feedback and the bolt driver. The bolt driver includes the ERob 70I rotary actuator \cite{zero_err_2025} and an octagonal socket;
    \item two pipes joined using a bolted chain clamp.
\end{itemize}

The remote site of the experimental setup is shown in Figure~\ref{fig:experimental-setup}.
The bolted joint used in the experiments is a chain clamp (Figure~\ref{fig:nut-bolt-closeup}), a chain-type collar placed around the two pipe ends and closed by tightening a single bolt.
The nut is captive in the clamp link and is therefore mechanically constrained by design; the clamp itself closes the reaction path during tightening.
The robot therefore only needs to engage and rotate the bolt; a second manipulator or an additional holding device is not required.
This arrangement is representative of the joining method documented for pipe flanges in IFMIF-DONES \cite{Ferre_2024_Design}, where the joint of pipe flanges is done using chain clamps to avoid cutting and welding for the replacement of pipe sections \cite{redondo_design_2025}, and bolting connections are among the minimum remote-handling activities declared for the facility \cite{Miccich__2025}.
From a network communication standpoint, all reported experiments were conducted over this local, PREEMPT\_RT-patched Ethernet network, a low-latency condition --- representative of the $\approx\SI{1}{\milli\second}$ telemanipulation latency budget specified for IFMIF-DONES \cite{Vazquez_2026} --- under which the haptic bilateral controller (Section~\ref{sec:manual-control}) renders force feedback transparently, without dedicated delay compensation.

It is worth highlighting that all trials involving direct teleoperation reported in Sections~\ref{sec:handling-of-visual-system-failures}--\ref{sec:handling-of-bolt-driver-failure} were performed by the same experienced operator, who had prior experience with the haptic device and completed preliminary training beforehand \cite{pagliara_human-robot_2023}.
Such a training contributes to mitigating learning effects during the actual experimental session.

\subsection{Prevention of interaction faults}\label{sec:prevention-of-interaction-faults}

To quantitatively assess the effectiveness of active compliance in preventing interaction faults, we performed 40 coupling experiments under conditions of misalignment between the octagonal socket and the bolt.
The coupling step (Section~\ref{sec:bolt-tightening-pipeline}, step 3) was executed 20 times with active admittance control (Scenario A) and 20 times with a rigid motion controller (Scenario B).
For each trial, we commanded a pre-scripted trajectory and recorded the normal interaction force during the collision phase between the socket and the bolt.

\begin{figure}
    \centering
    \includegraphics[width=\columnwidth]{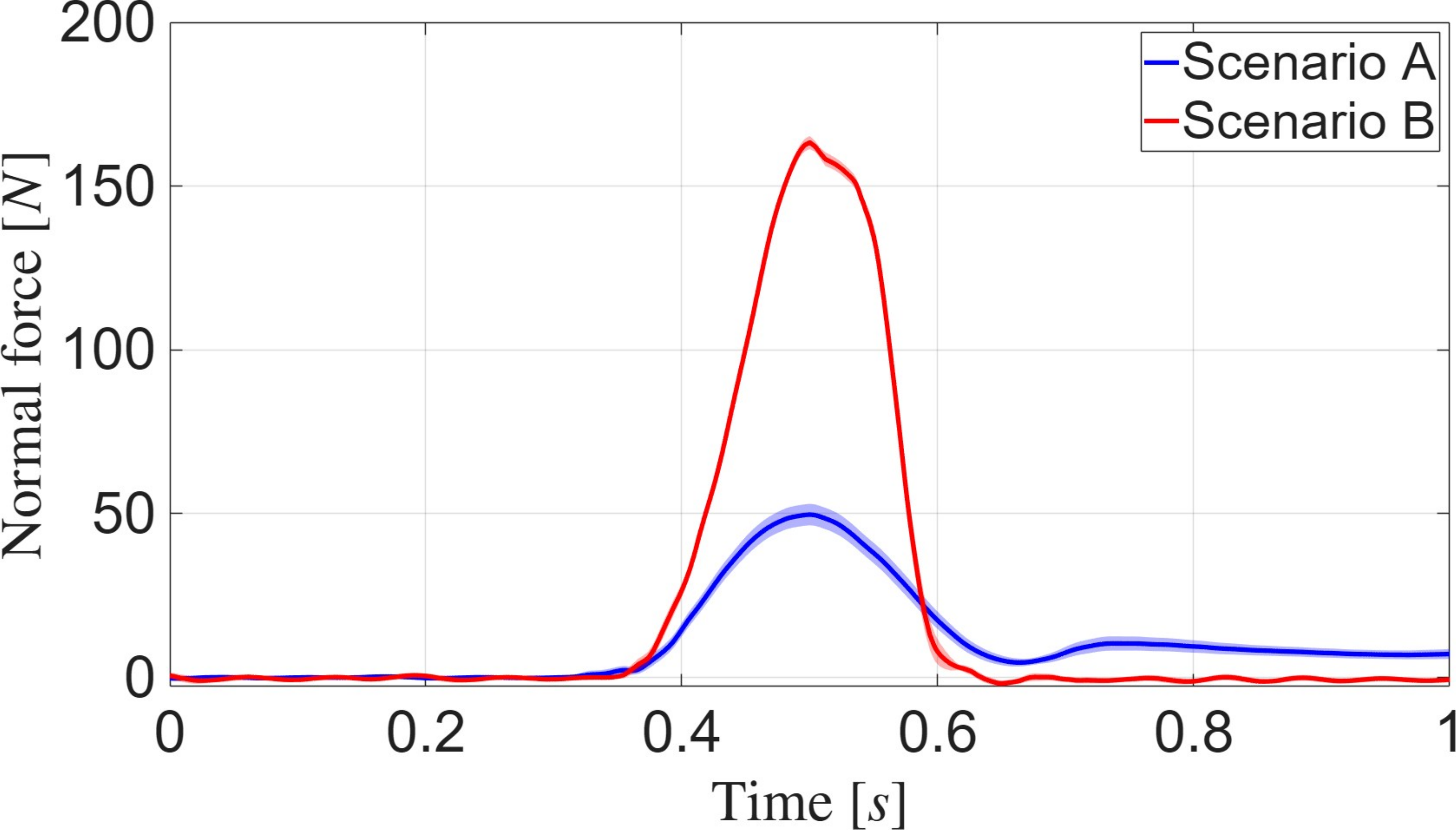}
    \caption{Comparison between the mean of the interaction force in Scenario A and Scenario B.}
    \label{fig:normal-force-comparison}
\end{figure}

Figure~\ref{fig:normal-force-comparison} reports mean and standard deviation of normal forces for both scenarios.
Active compliance reduced the peak force from \SIrange[range-units=single, range-phrase=$\pm$]{163.3}{4.1}{\newton} to \SIrange[range-units=single, range-phrase=$\pm$]{49.7}{11.0}{\newton} (\SI{69.57}{\percent} reduction, Welch's t-test $p = \num{5.7e-11}$), and the mean contact force from \SIrange[range-units=single, range-phrase=$\pm$]{43.6}{8.0}{\newton} to \SIrange[range-units=single, range-phrase=$\pm$]{13.3}{3.7}{\newton} (\SI{69.49}{\percent} reduction, $p = \num{6.7e-8}$).
The reduction is also significant under a non-parametric Mann-Whitney U test ($p < 0.001$ for both metrics), with large effect sizes (Cohen's $d > 10$ and $d > 5$, respectively).

Furthermore, active compliance reduced the trial-to-trial variability of the mean contact force (SD \SI{3.7}{\newton} vs \SI{8.0}{\newton}, ratio $0.46$, \SI{95}{\percent} bootstrap CI $[0.25, 0.81]$), making the interaction behavior more repeatable.
A Brown-Forsythe test confirms the variance reduction ($p = 0.04$).
For peak force, the dispersion did not differ significantly between scenarios (ratio $2.68$, \SI{95}{\percent} CI $[0.29, 7.41]$, $p = 0.21$).

In all rigid-control trials (\SI{100}{\percent}) the safety mechanism triggered due to excessive interaction forces, requiring manual reset.
In contrast, no safety stop was triggered in the active-compliance trials (\SI{0}{\percent}).
These results demonstrate that active compliance not only prevents force-triggered stops but also substantially reduces both the magnitude and the variability of interaction forces, improving the repeatability and safety of the coupling operation.

\subsection{Handling of incorrect bolt pose estimates}\label{sec:handling-of-visual-system-failures}

To perform a formal assessment of the system's capability to support the human operator in the detection and management of faults in the external autonomous system, we execute steps 2 and 3 of the procedure described in Section~\ref{sec:bolt-tightening-pipeline}, simulating a bolt-pose estimate error of \SI{0.0426}{\meter} in translation.
In such trials, we assume that the human operator fails to detect the estimation fault during the validation phase of step 2 and proceeds with the execution of step 3.
The fault is detected only during the execution of step 3, through the visual feedback provided by the camera, leading the operator to stop the automatic execution and take manual control.
The interruptions are deliberately induced to exercise the contingency-handling mechanism; their frequency is therefore a property of the experimental design and of the external perception system, which is emulated in this work.
The validation focuses on the ability to detect, stop, and recover from the contingency once it has arisen.
The validation is considered successful if the operator is able to complete the coupling with the bolt via direct teleoperation of the FR.

\begin{figure}
    \centering
    \includegraphics[width=\columnwidth]{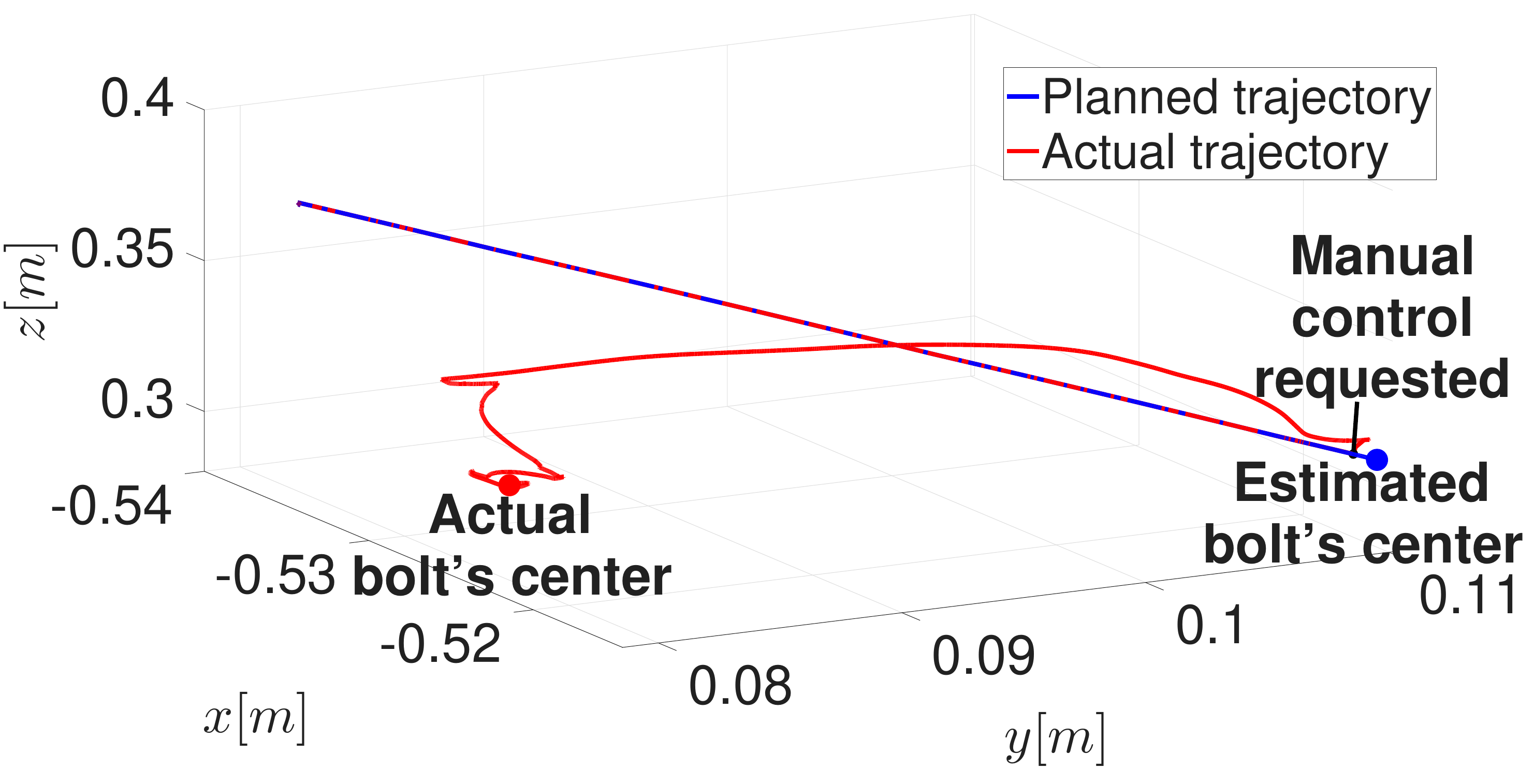}
    \caption{Comparison between the planned coupling trajectory and the one actually followed by the FR after human intervention.}
    \label{fig:coupling-trajectories-comparison}
\end{figure}

Figure~\ref{fig:coupling-trajectories-comparison} illustrates both the planned trajectory generated by the motion planner and the actual trajectory followed by the robot.
The results showcase that, once the fault is detected, the human operator is able to take manual control, and complete the coupling task.
These observations confirm that the system provides all the necessary tools to enable the human operator to detect faults and effectively manage contingencies.
However, they do not offer any insight into the effectiveness of the operator's intervention in completing the operation manually.
Therefore, in the accompanying video we provide a qualitative assessment of the latter.

\subsection{Handling of bolt driver failures}\label{sec:handling-of-bolt-driver-failure}

Since the proposed system aims at ensuring manual contingency handling through direct teleoperation, we evaluate the direct teleoperation system's capability to handle a bolt driver failure, by letting the human operator manually perform the tightening task.

The task consists of alternating between two steps, i.e., step 3 and step 4 of the operation described in Section~\ref{sec:bolt-tightening-pipeline}, executed by the human operator via the HD. This alternation is necessary due to joint limits of the FR and the risk of self-collisions, which prevent the bolt from being fully tightened in a single step. 
The operation is considered successful if the bolt is correctly tightened.
If at any step the force exerted exceeds a safety threshold, causing the robot to stop, or if the FR undergoes a self-collision, the operation is considered failed.
Throughout the execution, the human operator can rely on both haptic and visual feedback to complete the coupling with the bolt, and on the real-time display of torque values to determine when the tightening has been successfully completed.

We perform 20 trials on the same day.
The results show a success rate of $95\%$, with a single failure due to self-collision of the FR, of which the operator remained unaware.
These results highlight the system's effectiveness in completing a complex, contact-rich task, while also suggesting that avoiding self-collision during manual recovery might depend on the operator's situational awareness and task-specific skill.
While formal conclusions on the latter shall be drawn from a multi-operator statistical study, which we will conduct in future works (Section~\ref{sec:future-works}), we justify the former by stressing that the full robot posture is indeed shown in our HRI's GUI, hence self-collisions' occurrence probability can be reduced by taking care of task completion and FR state simultaneously.
The choice of 20 trials reflects a balance between obtaining a sufficiently large sample to assess the consistency of the recovery procedure and limiting operator fatigue as trials progress, which would have degraded performance for reasons unrelated to the system's inherent capabilities.
Consistently with established practice in teleoperated contact-rich tasks \cite{Boessenkool_2018, pagliara_safe_2024, pagliara_human-robot_2023}, this sample size is appropriate for demonstrating functional feasibility of the recovery mechanism.

\begin{figure}
    \centering
    \includegraphics[width=\columnwidth]{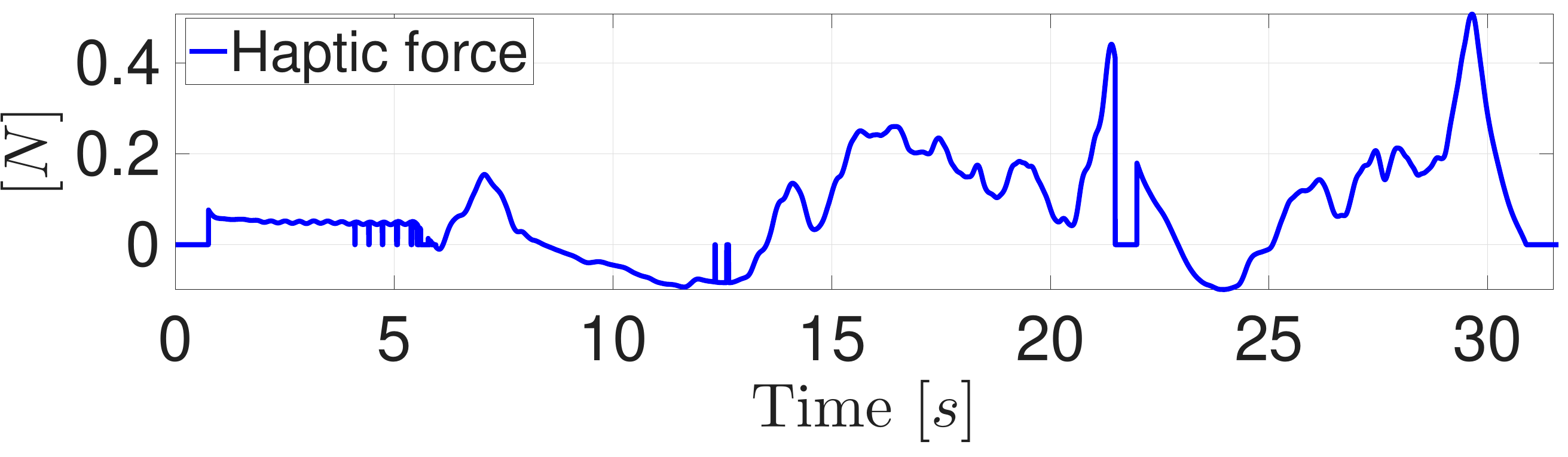}
    \caption{Forces rendered on the HD.}
    \label{fig:haptic-force}
\end{figure}

\begin{figure}
    \centering
    \includegraphics[width=\columnwidth]{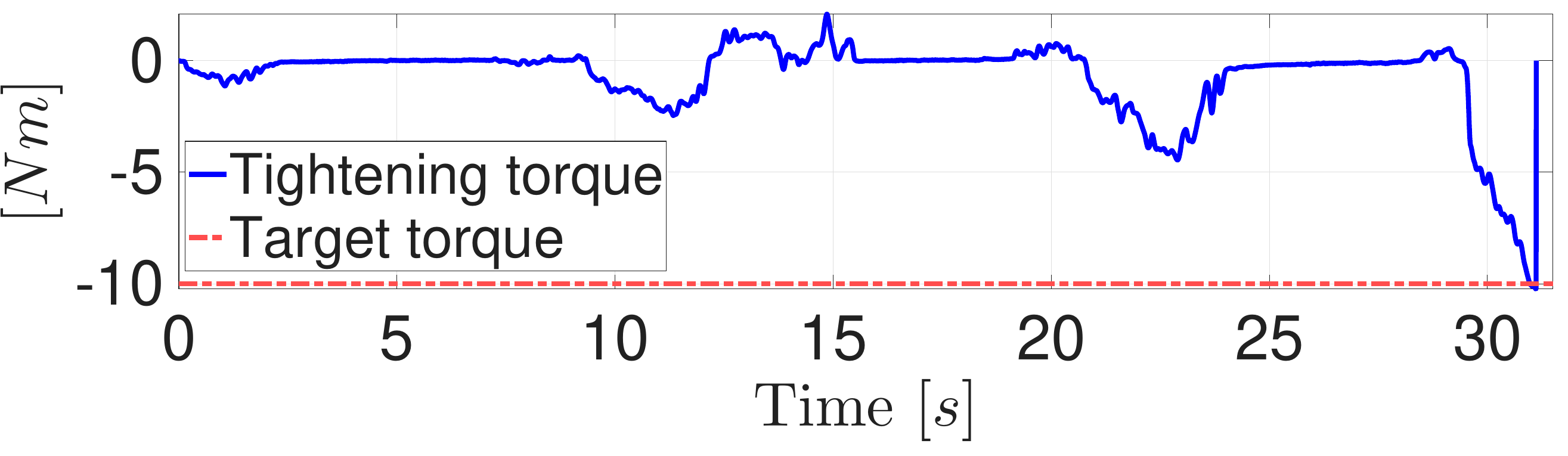}
    \caption{Tightening torque profile.}
    \label{fig:tightening-torque}
\end{figure}

A demonstration of a manual tightening execution is provided in the accompanying video.
For the sake of completeness, in this section we also report the outcomes of one successful trial, in terms of both the haptic feedback perceived by the human operator and the measured tightening torque.
The former is shown in Figure~\ref{fig:haptic-force}, while the latter is presented in Figure~\ref{fig:tightening-torque}.
Visual inspection of the haptic force profile reveals a slow dynamics, which enhances both operator comfort and situational awareness, facilitating the completion of the operation even in the presence of delays of the visual feedback.
The time required to assume manual control (from stopping the robot to effective engagement through the haptic device) has been recorded to be \SIrange[range-units=single, range-phrase=--]{3}{10}{\second} across the 20 trials, defining the overhead of the handover procedure.
This datum is specific to an experienced operator; automatic alignment between the haptic stylus and the follower socket would reduce it further, but this would come at the cost of upgrading the hardware and introducing an additional source of failure in the end-to-end system.

\section{Conclusions}\label{sec:conclusions}

\subsection{Discussion}\label{sec:discussion}

This paper proposed a control framework, and a relative system, for dependable robotized bolting operations. 
The system features a middle/low-level control that guarantees accurate control of the bolt driver tip and active compliance throughout the whole operation, ensuring a safe interaction also in case of fault conditions.

To enhance the human operator's situational awareness and ensure a smooth transition to manual control in case of contingencies, we designed a multimodal HRI that includes real-time visualization of all remote site state information, including camera images, as well as several input interfaces to support different control modes. 
We also ensured the proper execution of the operation by designing a high-level SV that coordinates the operation flow and the control mode switching, ensuring consistency with the SVC paradigm in which the human operator's authority is always preserved. 

The experimental validation of the system on a representative bolting task involving pipe flanges joining demonstrated the system's capability to effectively manage bolting operations, enabling reliable detection of the main system faults and operator intervention to handle them. 
However, it also suggested that the residual risk of self-collision during manual recovery is tied to the operator's situational awareness and task-specific skill in using the full-posture feedback already available through the HRI, rather than to a hardware limitation: we plan to validate this hypothesis directly in the multi-operator study proposed in Section~\ref{sec:future-works}.

Following the scope and applicability envelope of the framework established in Section~\ref{sec:scope-and-applicability}, and the fault-class coverage stated in Section~\ref{sec:experimental-validation}, the remainder of this discussion builds on that basis and follows a single arc: we first position the framework relative to the closest autonomous, teleoperated, and manual alternatives in Section~\ref{sec:comparison-with-other-methods}; we then examine the implementation- and hardware-specific limitations of the current demonstrator in Section~\ref{sec:limitations}; we discuss the failure modes of the teleoperated recovery mechanism itself, as observed across the trials, in Section~\ref{sec:teleoperation-failure-modes}; we close with Section~\ref{sec:technological-readiness-level} by justifying, on this basis, the TRL-5 claimed in Section~\ref{sec:introduction}.

\subsubsection{Comparison with other methods}
\label{sec:comparison-with-other-methods}

\begin{table*}
\centering
\caption{Qualitative comparison of execution conditions.}
\label{tab:timestudy}
\begin{tabularx}{\textwidth}{
    >{\hsize=0.8\hsize\raggedright\arraybackslash}X|
    >{\hsize=1.4\hsize\raggedright\arraybackslash}X|
    >{\hsize=1.3\hsize\raggedright\arraybackslash}X|
    >{\hsize=1.1\hsize\raggedright\arraybackslash}X|
    >{\hsize=0.4\hsize\raggedright\arraybackslash}X
}
\textbf{Condition} & \textbf{Nominal cost} & \textbf{Contingency cost} & \textbf{Operator occupancy} & \textbf{Source} \\
\hline\hline
Manual & $\approx$\SIrange[range-units=single, range-phrase=--]{339}{623}{\second} per 5-fastener sequence, including reading instructions/docs & --- & Continuous, on-site & \cite{Fau_2025} \\
\hline
Fully teleoperated & \SIrange[range-units=single, range-phrase=--]{163}{240}{\second} per bolt & \SIrange[range-units=single, range-phrase=--]{12}{635}{\second} per event; $\approx$30\% on overall task completion upon contingencies & Continuous, at controls & \cite{Boessenkool_2018} \\
\hline
Autonomous & $\approx$\SI{5}{\second}/operation, including $\approx$\SI{2.5}{\second} perception & 4/5 classes escalate w/o recovery path & 0\% nominal; undefined on failure, likely on-site & \cite{Manyar_2026,espinosa_peralta_robust_2023} \\
\hline
Proposed & Same as autonomous & Same as teleoperated, contingent fraction only & $\mathbb P(\text{contingency})$ $\times$ occupancy/event & ours \\
\end{tabularx}
\end{table*}

Our framework is not proposed as an alternative to autonomous bolting, but as a complementary recovery layer for contingencies that autonomous systems in this domain are documented not to resolve on their own.
Consistently with this scope, the experimental campaign of Section~\ref{sec:experimental-validation} validates precisely these fault-detection and contingency-handling capabilities; end-to-end metrics, such as task success rate, are excluded in our study because they would include, as noted in Section~\ref{sec:methodology}, the coupled autonomous executor and its perception subsystem, which lie outside the present scope.
The closest autonomous baseline in the literature, a decision-tree-based failure detector validated on ten real industrial parts under hole-pose uncertainties of up to \SI{4}{\milli\meter}/\SI{3}{\degree} \cite{Manyar_2026}, illustrates why: it reaches perfect precision, recall and F1-score on its own real-world test set, yet of the five failure modes it detects, only a timeout-based mode triggers an autonomous reattempt --- the remaining four, the interaction and alignment faults closest to those we address in Sections~\ref{sec:prevention-of-interaction-faults}--\ref{sec:handling-of-visual-system-failures}, trigger a call for human assistance, with no autonomous recovery and no interface specified for the operator to act on it.
To the best of the authors' knowledge, no published autonomous fastening system both detects and recovers from these classes; compared with this and other baselines, the proposed framework integrates exactly this missing piece --- a general autonomy-failure contingency mechanism --- rather than a faster or more complete autonomous executor.

Relative to full teleoperation, the time saved with the proposed approach follows from the nominal phase executing autonomously along a planned, collision-aware trajectory toward the target pose, assumed known (Section~\ref{sec:methodology}), rather than from the teleoperated recovery itself being faster.
In the only real-facility, task-comparable time-action study available --- direct teleoperation of tile-carrier installation at JET, itself a bolt-running operation --- median completion time was \SIrange[range-units=single, range-phrase=--]{163}{240}{\second} per bolt depending on operator experience, and approximately \SI{30}{\percent} of total execution time was attributable to non-nominal actions, with individual delays ranging from \SI{12}{\second} to over \SI{10}{\minute}, mostly caused by small misalignments \cite{Boessenkool_2018}.
This is the cost a fully teleoperated system pays on every operation; the proposed framework instead pays it only on the contingent fraction, so the time saved scales with how large the nominal fraction is.
Because the operator intervenes only during this contingent fraction rather than continuously, a single operator could, in principle, supervise several autonomous instances of the system in parallel, stepping in only where a contingency arises: an operating mode that full teleoperation, requiring continuous engagement, does not afford.
Against fully manual execution, even with a calibrated torque wrench, completion time has been reported to be dominated by instruction handling and manual documentation rather than by tightening itself \cite{Fau_2025}; the advantage claimed over manual execution is reduced continuous labor and site exposure, not per-operation speed.

We define end-to-end performance as the nominal execution time plus the sum of detection, intervention, recovery and re-validation time, at the driving torque required by the facility's post-maintenance acceptance testing --- a quality criterion prescribed by the target facility rather than one we impose \cite{Miccich__2025} --- summarized qualitatively in Table~\ref{tab:timestudy}.
Operator occupancy under this definition is proportional to the probability that a given operation raises a contingency: an operation that completes nominally claims no operator's time, while one that does not claims \SIrange[range-units=single, range-phrase=--]{12}{635}{\second}, of which $\approx\SI{30}{\percent}$ \cite{Boessenkool_2018} is taken performing non-nominal operations.
We only report this figure from the comparable JET bolt-running task because a specific contingency rate for our target facility is not available.
Indeed, valuing, in practical terms, the actual time gain achieved by our framework, however, depends on both the operator's ability at the specific recovery action on its own, and on the coupled autonomous system's performance \cite{Perkins_2025}.

On workload, the closest quantitative analogue we found --- force-feedback teleoperation in hazardous-environment maintenance --- reports operator workload at parity with hands-on manual work when haptic feedback is used \cite{Lessard_1995}; since our framework engages the operator only during the contingent fraction of the task, this parity implies that the operator's aggregate workload falls with the size of the nominal fraction, in line with the occupancy argument above.
An instrumented workload study on our own system is left as future work.
Finally, removing the supervisory and HRI layer from the present system leaves detection with no specified path to recovery, exactly as observed for four of the five failure modes in \cite{Manyar_2026} and consistent with the undetected self-collision reported in Section~\ref{sec:handling-of-bolt-driver-failure}.

\subsubsection{Implementation limitations}\label{sec:limitations}

The limitations discussed below are specific to the current hardware and software instantiation, not to the supervisory framework itself (the latter are discussed in Section~\ref{sec:scope-and-applicability}).
For each, we identify which block of the framework (Figure~\ref{fig:SVCS}) would accommodate the corresponding upgrade.

\begin{itemize}
\item Unconstrained fastener (FR block).
Where the bolt's counterpart is not mechanically constrained, as it is in the chain-clamp arrangement used here (Section~\ref{sec:experimental-setup}), a reaction-torque feature integrated in the end-effector \cite{Tao_2022, Ferre_2024_Forces} or high-torque passive compliance in the tool \cite{Noguchi_2021} would extend the FR block to cover this case, without requiring changes to the supervisory logic.
A configuration in which the counterpart must itself be actively held is a distinct, framework-level limitation, discussed in Section~\ref{sec:scope-and-applicability}.
\item Fastener size (FR block).
Dimensions outside the range supported by the current fixed-socket end-effector cannot be handled without upgrading it. Modular \cite{Li_2022} or reconfigurable \cite{Abhishek_2023} end-effectors, with a width-adjustable screwing mouth covering a range of fastener widths \cite{Tao_2022}, would extend the FR block as a direct replacement, again without modifying the supervisory logic.
\item Communication delay (manual control block).
As specified in Section~\ref{sec:experimental-setup}, the teleoperation layer was operated within a low-latency, PREEMPT\_RT-patched network \cite{Reghenzani_2019}, representative of the $\approx\SI{1}{\milli\second}$ telemanipulation budget specified and met by IFMIF-DONES's own infrastructure \cite{Vazquez_2026}, lying orders of magnitude below the multi-second delays typical of orbital/planetary telemanipulation \cite{Panzirsch_2022,Panzirsch_2024}.
Beyond this envelope, delay compensation becomes necessary; the choice of the specific technique is orthogonal to the proposed framework, since established methods such as the Time-Domain Passivity Approach \cite{Panzirsch_2022} or model-augmented telemanipulation \cite{Panzirsch_2024} can be integrated into the teleoperation layer without modifying our supervisor, switching conditions, or handover mechanisms: communication delays represent a general constraint of human-oriented haptic teleoperation \cite{Hirche_2012}, not specific to this implementation.
\item Workspace geometry (automatic control / manual control blocks).
Geometries that exceed the current manipulation or collision-awareness capability, including complex workspaces with multiple obstacles or tightly constrained access, are not addressed by the present implementation; extending the validation to such scenarios is listed as future work in Section~\ref{sec:future-works}, and, as with the limitations above, would extend the capability of the automatic- and manual-control blocks without requiring changes to the supervisory logic.
\item Radiation hardening (FR block).
Radiation-induced degradation of the tool and sensing components --- a relevant concern for deployment in nuclear facilities --- has not been assessed.
Hardening the specific hardware against radiation, as discussed for comparable bolting tools in \cite{Noguchi_2021}, is an engineering concern orthogonal to the supervisory framework itself.
\end{itemize}

Taken together, these limitations belong to the hardware and control layers of the instantiated system rather than to the supervisory framework, which is designed to accommodate the corresponding upgrades without requiring modification to its logic.
The extent to which the experimental campaign itself was scoped --- fastener size, tool-changing, and multi-bolt sequencing --- is stated in Section~\ref{sec:experimental-validation}.

\subsubsection{Teleoperation failure modes}
\label{sec:teleoperation-failure-modes}

Our teleoperated intervention is a recovery mechanism that, like the autonomous execution it complements, has its own failure modes.
We identify these failure modes explicitly to clarify the scope of the framework and to guide future improvements.
Table~\ref{tab:teleoperation-failure-modes} summarizes the principal failure modes, their causes, how they become observable, their mitigation, and whether the current system covers them.

\begin{table*}
    \centering
    \caption{Teleoperation failure modes}
    \label{tab:teleoperation-failure-modes}
    \begin{tabularx}{\textwidth}{
        >{\hsize=0.9\hsize\raggedright\arraybackslash}X|
        >{\hsize=1.1\hsize\raggedright\arraybackslash}X|
        >{\hsize=1.1\hsize\raggedright\arraybackslash}X|
        >{\hsize=1.2\hsize\raggedright\arraybackslash}X|
        >{\hsize=0.7\hsize\raggedright\arraybackslash}X
    }
        \textbf{Failure mode} & \textbf{Cause} & \textbf{How it becomes observable} & \textbf{Mitigation in current system} & \textbf{Covered?} \\
        \hline
        \hline
        Excessive interaction force	& Operator commands a pose that would cause high contact force & Force/torque sensor reading; admittance controller limits force & Safety stop triggers when force exceeds threshold & Yes (Section~\ref{sec:prevention-of-interaction-faults}) \\
        \hline
        Self-collision & Operator unaware of robot's joint configuration or external geometry & Visual feedback; GUI state display	& Operator monitors via GUI; safety stop if collision occurs & Yes (one failure occurred in Section~\ref{sec:handling-of-bolt-driver-failure}) \\
        \hline
        Joint-limit violation & Operator commands a pose outside the robot's kinematic envelope & GUI joint configuration display; robot stops & Operator monitors via GUI; controller stops & Yes \\
        \hline
        Inability to complete tightening in a single step & Joint limits and self-collision risk prevent full rotation & Operator observes partial completion & Procedure split into alternating coupling/tightening steps & Yes (Section~\ref{sec:handling-of-bolt-driver-failure}) \\
        \hline
        Communication delay	& Network latency; visual feedback lag & Delayed response to commands & PREEMPT\_RT patch \cite{Reghenzani_2019}; visual feedback lag is accepted as a known limitation & Partially (patched OS as in Section~\ref{sec:experimental-setup})
    \end{tabularx}
\end{table*}

The two failure modes observed in our experiments were excessive interaction force (triggering the safety threshold in \SI{0}{\percent} of active-compliance trials, as reported in Section~\ref{sec:prevention-of-interaction-faults}, and in \SI{100}{\percent} of rigid-control trials when the operator exceeded the threshold) and self-collision (one failure out of 20 manual trials, reported in Section~\ref{sec:handling-of-bolt-driver-failure}).
The latter highlights that avoiding self-collision during manual recovery depends on the operator making effective use of the robot state displayed in the HRI's GUI: we expect this capability to scale with operator training and task-specific skill rather than with additional feedback channels.
This observation motivates the multi-operator study suggested in Section~\ref{sec:future-works}, aimed at characterizing how operator skill affects the incidence of this failure mode.

It is therefore worth noting that the teleoperated intervention is not claimed to be failure-free; rather, it is a recovery mechanism that, when it fails, leaves the system in a safe state (stopped, with the operator informed), from which the operation can be reattempted or the task re-planned.
This is consistent with the SVC paradigm \cite{sheridan_telerobotics_1992}, in which the human operator retains ultimate authority and can respond to failures of the recovery itself.

As established in Section~\ref{sec:experimental-setup}, our 20 manual trials were conducted by a single experienced operator.
The observed \SI{95}{\percent} success rate is therefore presented as evidence of the functional feasibility of the recovery mechanism for the tested operator, rather than as a broadly generalizable population statistic.
This framing is consistent with common practice in teleoperation system studies \cite{Parasuraman_2000, pagliara_human-robot_2023, pagliara_haptic_2025}, where the primary object of evaluation is the architecture itself rather than human performance variability.
The operator's prior experience and preliminary training (Section~\ref{sec:experimental-setup}) mitigated learning effects during the 20 trials, consistently with recommended practice for a realistic deployment \cite{pagliara_human-robot_2023}.
The single failure that occurred is consistent with self-collision occurrence being, at least in part, an operator-dependent outcome rather than a fatigue effect.
Therefore, we conclude that
\begin{enumerate}[label=(\roman*)]
\item the binary success outcome is primarily determined by the robot's kinematic envelope, collision-awareness capabilities, and available feedback modalities, all of which are implemented in the control system;
\item the operator's ability does influence motion quality and duration, and might benefit from further task-specific training to diminish the probability of self-collisions.
\end{enumerate}

\subsubsection{Technological Readiness Level}
\label{sec:technological-readiness-level}

We justify the TRL-5 claim against the nine-level scale adopted for Horizon Europe \cite{TRL}, which distinguishes TRL-4 ("technology validated in laboratory environment"), TRL-5 ("technology validated in relevant environment") and TRL-6 ("technology demonstrated in relevant environment").
The step beyond TRL-4 is on the environment, not on the integration: the admittance controller, torque control, multimodal HRI, haptic teleoperation and supervisory logic were already exercised together as one system, which a purely laboratory validation could equally have shown.
What makes the environment relevant rather than generic is that the task, the hardware, and the fault classes are the facility's own: the chain-clamp arrangement (Section~\ref{sec:experimental-setup}) is documented for pipe-flange joining in IFMIF-DONES-class facilities independently of this work \cite{Ferre_2024_Design,redondo_design_2025}; the injected faults are drawn from the facility's documented minimum remote-handling task set and failure modes \cite{Miccich__2025,Valenzuela_2023}; and the teleoperation channel was operated within the facility's own latency budget \cite{Vazquez_2026}, as reported in Section~\ref{sec:experimental-setup}, not under an incidentally convenient local-network delay.

TRL-6 is not yet reached, hence we state the gap as follows. As detailed in Section~\ref{sec:scope-and-applicability}, perception is emulated rather than integrated; the validation covers a single fastener size, tool and operator (Sections~\ref{sec:experimental-validation}, \ref{sec:experimental-setup} and \ref{sec:limitations}--\ref{sec:teleoperation-failure-modes}), so the operator variability documented in comparable facility studies \cite{Boessenkool_2018} is not represented; and, as also reported in Section~\ref{sec:scope-and-applicability}, the facility's own post-maintenance acceptance test --- instrumentation check, leak test --- was not performed \cite{Miccich__2025}, nor was the hardware exposed to radiation (Section~\ref{sec:limitations}) or to the facility's physical access constraints.
Each of these marks a concrete step toward TRL-6.

We therefore claim TRL-5: the framework and its demonstrator are validated, not yet demonstrated, in an environment representative of the target facility.
The claim applies to this integrated demonstrator, not to the general problem of fully autonomous robotized bolting.

\subsection{Future works}\label{sec:future-works}

Based on the experimental findings of this paper, in future work we will:
\begin{enumerate*}[label=(\arabic*)]
\item integrate a robust vision system for bolt pose estimation;
\item extend the validation to different bolting scenarios including the presence of obstacles and uncertainty factors, addressing the workspace-geometry limitation noted in Section~\ref{sec:limitations};
\item integrate an autonomous anomaly-detection module, building on existing force-signal classifiers \cite{Manyar_2026}, to provide quantitative detection-time and detection-accuracy figures --- metrics that, to the best of the authors' knowledge, are seldom reported in the existing fastening fault-detection literature --- and enable a fully self-contained end-to-end evaluation;
\item conduct an instrumented multi-operator study (e.g., with NASA-TLX and performance metrics) as done in other works \cite{Boessenkool_2018, Fauquet_Alekhine_2024}.
\end{enumerate*}

\printcredits

\section*{Declaration of competing interest}
The authors declare that they have no known competing financial interests or personal relationships that could have appeared to influence the work reported in this paper.

\section*{Data availability}
Data will be made available on request.

\bibliographystyle{elsarticle-num}

\bibliography{bibliography}

\end{document}